\documentclass[letterpaper, 10 pt, conference]{ieeeconf}  

\IEEEoverridecommandlockouts                              

\usepackage{graphics} 
\usepackage{epsfig} 
\usepackage{times} 
\usepackage{amsmath} 
\usepackage{amssymb}  
\usepackage{soul}
\usepackage{graphicx}
\usepackage{algorithm} 
\usepackage{algpseudocode} 
\usepackage{multirow} 

\usepackage{makecell}
\usepackage{booktabs}
\usepackage{cuted} 
\usepackage{caption} 
\newcommand{\argmin}{\operatornamewithlimits{argmin}}

\usepackage{xcolor}
\usepackage{multicol}
\definecolor{mycitecolor}{RGB}{71, 191, 38}
\definecolor{mylinkcolor}{RGB}{40, 115, 201}
\definecolor{DGreen}{RGB}{107, 190, 35}
\definecolor{darkblue}{rgb}{0.1, 0.1, 1}

\usepackage[skip=0pt]{caption}
\makeatletter
\let\NAT@parse\undefined
\makeatother
\usepackage[colorlinks=true, citecolor=mycitecolor,linkcolor=mylinkcolor,urlcolor=mycitecolor]{hyperref}

\title{\LARGE \bf
UDON: Uncertainty-weighted Distributed Optimization for Multi-Robot Neural Implicit Mapping under Extreme Communication Constraints
}

\author{Hongrui Zhao$^{1*}$, Xunlan Zhou$^{2,3*}$, Boris Ivanovic$^{4}$, Negar Mehr$^{5}$%
 \thanks{$^{1}$ Department of Aerospace Engineering, University of Illinois Urbana-Champaign
         {\tt\small hongrui5@illinois.edu}}%
 \thanks{$^{2}$ School of Intelligent Science and Technology, Nanjing University
         {\tt\small wyattzhouxl@smail.nju.edu.cn}}%
 \thanks{$^{3}$ National Key Laboratory for Novel Software Technology,
 Nanjing University}%
 \thanks{$^{4}$ NVIDIA Research
         {\tt\small bivanovic@nvidia.com}}%
 \thanks{$^{5}$ Department of Mechanical Engineering, University of California, Berkeley
         {\tt\small negar@berkeley.edu}}%
 \thanks{$^{*}$ These authors contributed equally to this work.}%
}

\begin{document}

\maketitle

\begin{strip}
  \begin{minipage}{\textwidth}\centering
    \vspace{-65pt}
        \includegraphics[trim={0cm 0cm 0cm 0cm},clip,width=0.70\columnwidth]{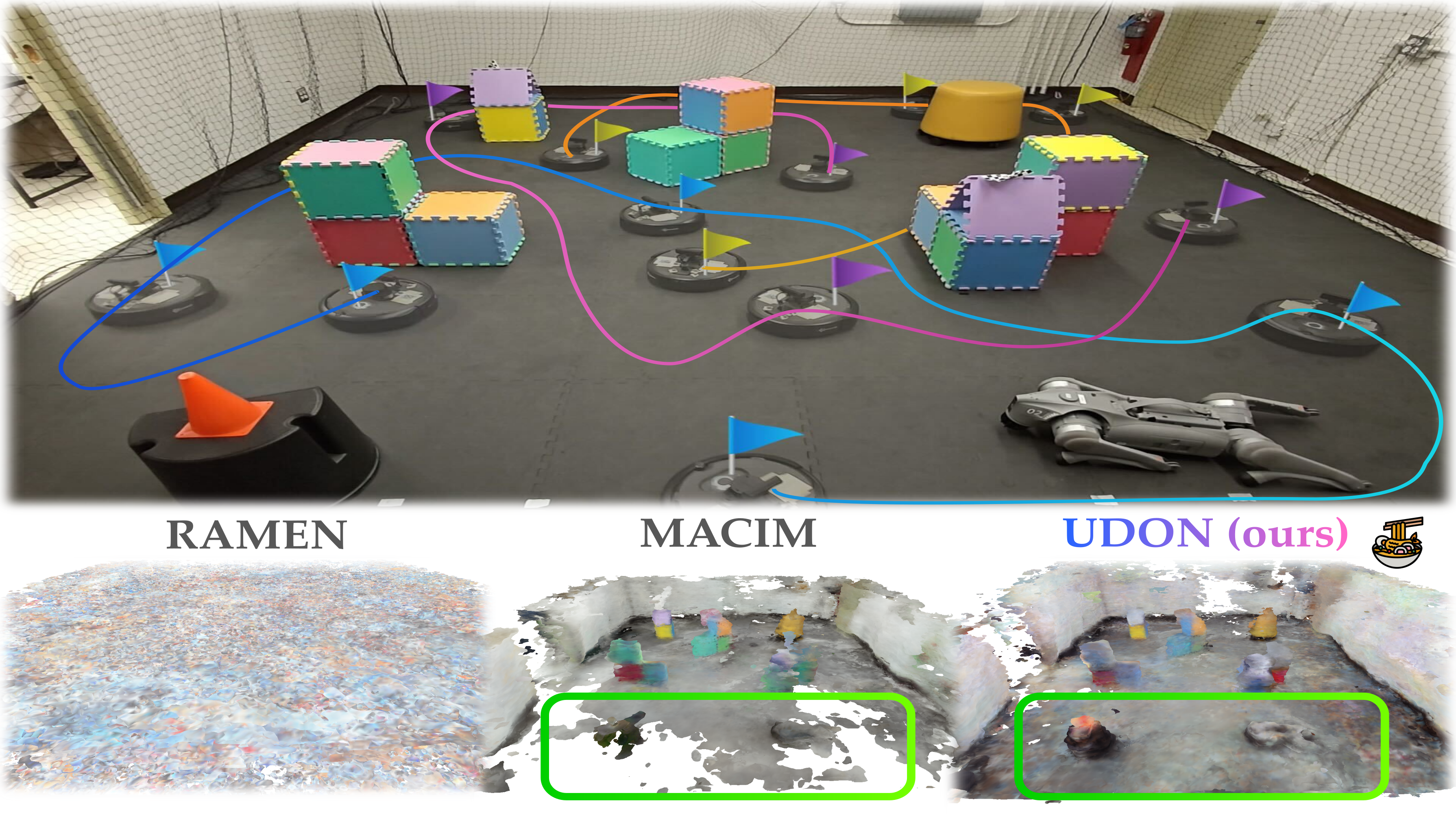}  
    \captionof{figure}{ 
    \textbf{UDON outperforms existing baselines in a three-robot mapping experiment conducted with a challenging 5\% communication success rate}.
    We conducted the experiment using a dataset collected with TurtleBot robots. 
    For visualization purposes, the trajectory of each agent is color-coded.
    While the baseline methods either fail to converge or produce incomplete reconstructions, UDON, in contrast, successfully fuses information from all agents to create a complete map of the scene.
    }
    \label{fig:teaser}
  \end{minipage}
\end{strip}


\begin{abstract}
Multi-robot mapping with neural implicit representations enables the compact reconstruction of complex environments. 
However, it demands robustness against communication challenges like packet loss and limited bandwidth. 
While prior works have introduced various mechanisms to mitigate communication disruptions, performance degradation still occurs under extremely low communication success rates. 
This paper presents UDON, a real-time multi-agent neural implicit mapping framework that introduces a novel uncertainty-weighted distributed optimization to achieve high-quality mapping under severe communication deterioration. 
The uncertainty weighting prioritizes more reliable portions of the map, while the distributed optimization isolates and penalizes mapping disagreement between individual pairs of communicating agents.
We conduct extensive experiments on standard benchmark datasets and real-world robot hardware.
We demonstrate that UDON significantly outperforms existing baselines, maintaining high-fidelity reconstructions and consistent scene representations even under extreme communication degradation (as low as 1\% success rate). 
The codes can be found at \href{https://iconlab.negarmehr.com/UDON/}{https://iconlab.negarmehr.com/UDON/}.
\end{abstract}

\section{INTRODUCTION}

Multi-robot systems can map large-scale environments more efficiently by exchanging local observations to build a unified scene representation.
A particularly promising recent approach is to use neural implicit scene representations, such as Neural Radiance Fields (NeRFs) \cite{NeRF}, which encode complex 3D geometry and appearance into a small amount of learnable parameters. 
Neural implicit maps are compact and can readily incorporate semantic information \cite{Inplace, Lerf}, making them suitable for multi-agent mapping applications ranging from planetary exploration \cite{Nebula, LunarCADRE} to multi-UAV survey missions \cite{UAV}. 
However, the performance of these collaborative systems is vulnerable to communication disruptions, such as limited bandwidth or environmental factors. 
To this end, this paper addresses a key challenge: enhancing the robustness of multi-agent neural implicit mapping against severe bandwidth limitations and packet loss.

While explicit representations can be fused through simple additive operations—for instance, by combining multiple point clouds—neural implicit maps require more complex strategies
One common strategy is to pose map fusion as a distributed optimization problem and drive local neural maps toward a \emph{consensus} using methods such as decentralized consensus alternating direction method of multipliers (C-ADMM) \cite{DiNNO,Di-NeRF}. 
This consensus yields per-agent replicas of a shared neural implicit map that encodes the team’s collective observations. 
In practice, however, bandwidth limits, latency, and packet drops render communication inherently asynchronous.
Under these conditions, the underlying distributed optimization may diverge and agents may fail to reach consensus, leading to degraded reconstruction fidelity and inconsistent scene understanding across the team~\cite{ramen}. 

To mitigate this problem, state-of-the-art methods such as RAMEN~\cite{ramen} utilize an uncertainty-weighted C-ADMM algorithm to bias the consensus towards more reliable neural maps.
RAMEN achieves robust mapping performance under challenging communication disruptions. 
However, its performance still degrades under extreme communication conditions (e.g.,  less than 20\% communication success rate).
Other multi-agent neural implicit mapping methods like MACIM \cite{MACIM} and OpenMulti \cite{OpenMulti} attempt a different approach, replacing C-ADMM with a ``consistency loss". 
Despite offering greater robustness to communication failures, these methods are often limited in their ability to capture fine environmental details, especially under severe communication constraints.

To achieve robustness in extreme communication scenarios, we introduce UDON, a uncertainty-weighted distributed optimization for multi-agent neural implicit mapping. 
Our approach models a multi-robot system as a communication graph where agents are nodes and communication links between them are edges. 
Building upon the MESA framework \cite{MESA}, a key innovation of our approach is to assign an independent \emph{dual variable} to each communication edge. 
This formulation departs from the previous methods that consolidate information from all edges into a single dual variable.
At each communication step, agents exchange their maps and their associated uncertainties with their neighbors. 
Each edge's dual variable is then updated \emph{independently}, weighted by its uncertainty, to penalize divergence between that specific pair of maps. 
These updated dual variables are then incorporated into each agent's loss function, guiding the optimization of their local map toward a consensus.

Through comprehensive experiments, we show that UDON consistently delivers higher reconstruction quality than existing methods. 
We demonstrate that the advantage is most pronounced in highly-unreliable communication settings, where prior methods often suffer significant performance degradation or fail to converge.
Finally, we validate UDON's effectiveness in real-world experiments conducted on TurtleBot hardware.
The data collected by these robots features trajectory coverage patterns that significantly differ from standard, hand-held camera benchmarks, providing a more realistic and challenging evaluation of UDON's performance.

Our key contributions are threefold: 
\begin{enumerate}
\def\labelenumi{\arabic{enumi}.}
\item We propose UDON, an uncertainty-weighted multi-agent neural implicit mapping framework designed for extreme communication scenarios. 
\item We demonstrate that UDON significantly outperforms existing baselines through extensive experiments, especially under highly-unreliable communication.
\item We validate the performance of UDON using data collected from real-world robot experiments.
\end{enumerate}

\section{RELATED WORKS}\label{Related Works}

\textbf{Single-agent Neural Implicit Mapping}. 
Recent advances in neural implicit mapping have enabled high-fidelity reconstruction and camera tracking by leveraging implicit scene representations. 
For instance, iMAP \cite{iMAP} demonstrates that a multilayer perceptron (MLP) can serve as the sole scene representation in real-time SLAM, achieving plausible hole-filling via implicit priors. 
Building on this idea, NICE-SLAM \cite{NICE-SLAM} introduces a multi-resolution feature grid to accelerate training and capture local details, though at the cost of reduced hole-filling capability. 
Co-SLAM \cite{Co-SLAM} combines coordinate encoding with multi-resolution hash grids, enabling fast convergence, high-fidelity mapping, and efficient memory use.  
Recent works such as active neural mapping \cite{activeneural} and NARUTO \cite{naruto} also integrate uncertainty into the neural mapping process. 
Quantifying the uncertainty of each agent's neural implicit map is a key step, as it allows us to identify the most reliable regions of the map. 
Building on these insights, UDON augments the Co-SLAM single-agent framework with a frequency-based uncertainty model, adapting it for the multi-agent mapping tasks.

\textbf{Multi-agent Neural Implicit Mapping}.
Recent efforts extend neural implicit mapping to multi-robot settings in order to exploit multi-view perception and improve reconstruction efficiency. 
DiNNO \cite{DiNNO} proposes a method for distributed neural network optimization using C-ADMM to achieve parameter consensus, a framework that Di-NeRF \cite{Di-NeRF} extends to enable the distributed training of NeRFs.
Building on these distributed optimization strategies, several methods have been proposed to enhance robustness against communication failures. 
MACIM \cite{MACIM} and OpenMulti \cite{OpenMulti} employ priors to mitigate reconstruction artifacts in unobserved regions and use a consistency loss to ensure robust training during communication failures.
RAMEN \cite{ramen}  introduces frequency-based uncertainty into the consensus process, allowing robust mapping under asynchronous and unreliable communication.
However, these methods still remain vulnerable under extreme communication conditions (e.g., a communication success rate below 20\%).
Our method UDON enhances robustness against highly unreliable communication by introducing a novel distributed optimization technique.

\textbf{Asynchronous Distributed Optimization}. 
Methods that adapt the C-ADMM algorithm for asynchronous communication can potentially be applied to multi-agent mapping under communication failures.
Early efforts introduced barrier-based schemes, such as partial barrier and bounded delay conditions \cite{barrier1, barrier2}, which allow progress despite straggling or delayed workers. 
While these approaches alleviate synchronization bottlenecks in distributed training, they are primarily tailored for data-center or server-based applications, and are less suited for real-time robotic mapping tasks that require continuous updates. 
More recently, MESA \cite{MESA} reformulated C-ADMM for collaborative SLAM by introducing an edge-based separable optimization framework, permitting asynchronous communication between robots while enforcing consistency over shared variables. 
Its extension, iMESA \cite{imesa}, further adapts this idea for online operation, integrating incremental solvers to achieve real-time distributed localization under sparse and unreliable communication.
Building on these advances, UDON draws inspiration from the design of MESA, but adapts it for multi-agent neural implicit mapping. 
\vspace{-5 mm}
\section{Preliminaries} \label{Preliminaries}
In this section, we introduce the basic C-ADMM algorithm for solving the multi-agent mapping problem, which our method, UDON, improves upon.
We model a team of agents jointly mapping an environment as a communication graph $\mathcal{G}=(\mathcal{V}, \mathcal{E})$,
where the agents are nodes $i \in \mathcal{V}$, and the communication links between them are edges $(i,j) \in \mathcal{E}$.
We use a subscript $i$ to specify variables that belong to agent $i$.
For any agent $i$, we denote its set of communicating neighbors as $\mathcal{N}_i = \{ j  \in \mathcal{V} \mid (i,j) \in \mathcal{E} \}$.

We consider the mapping process over discrete time steps, or iterations, indexed by the superscript $t$.
Every agent continuously populates a local dataset, $\mathcal{R}_i^t$, with newly captured RGB-D images and their corresponding poses.
Each agent also maintains a local environment map, parameterized by $\Theta_i^t$.
At every mapping iteration $t$, the agents communicate by sharing their respective map parameters, $\Theta_i^t$, with their current neighbors in $\mathcal{N}_i^t$.
Following this exchange, each agent updates its map to balance two key objectives: ensuring a high-fidelity reconstruction of its local data $\mathcal{R}_i^t$ and maintaining consensus by aligning its map with those of its neighbors $\Theta_j^t$.

To achieve these objectives, we use the C-ADMM algorithm, where each agent alternates between a \emph{dual update} and a \emph{primal update} at each iteration \cite{DiNNO,Di-NeRF,ramen}.
First, the dual update adjusts a \emph{dual variable}, $p_i$, which accumulates the divergence between an agent's map and those of its neighbors over time. 
The update rule is:
\begin{equation}
    p_i^{t+1} = p_i^t + \rho \sum_{j \in \mathcal{N}_i^t} ( \Theta_i^t - \Theta_j^t),
    \label{eq:dual_update}
\end{equation}
where $\rho$ is a user-defined step size. 
Following this, the primal update solves for the optimal map parameters, $\Theta_i^{t+1}$, by minimizing an objective that balances local reconstruction accuracy with penalties for consensus violations:
\begin{align}
    \label{eq:primal_update}
    \Theta_i^{t+1} =& \argmin_{\Theta_i} \; L^{obj}_i(\Theta_i, \mathcal{R}_i^t) + \langle \Theta_i, p_i^t  \rangle \nonumber \\
    & + \rho \sum_{j \in N_i} \left \Vert  \Theta_i - \frac{\Theta_i^t + \Theta_j^t}{2} \right \Vert_2^2,
\end{align}
where $\langle \cdot, \cdot \rangle$ is the dot product, and $L^{obj}_i$ is the local objective function that measures reconstruction error against the private dataset $\mathcal{R}_i^t$. 
We solve this local minimization for agent $i$ by taking $B$ steps of gradient descent, where $B$ is a user-defined constant.


\section{Method} \label{Method}
In this section, we detail the three core components of the UDON framework: (1) the neural implicit map representation, (2) our method for quantifying map uncertainty, and (3) our uncertainty-weighted distributed optimization for map updates, which is an improvement upon the methods in Section \ref{Preliminaries}. 
A summary of the overall approach is provided in Algorithm \ref{procedure_UDON}.

\subsection{Neural Implicit Representation} \label{method:mapping}
In UDON, each agent $i$ adopts the neural implicit representation from Co-SLAM \cite{Co-SLAM} to model its local map of the scene.
UDON's scene representation comprises three components: a multi-resolution hash grid $V_{\Theta_i}$ \cite{instantNGP} that captures features across multiple levels of detail, a geometry MLP decoder $F_{\tau_i}$, and a color MLP decoder $F_{\phi_i}$.
Decoders $F_{\tau_i}$ and $F_{\phi_i}$ are pretrained on the the NICE-SLAM apartment dataset \cite{NICE-SLAM}, and we fix their parameters during mapping to simplify uncertainty computations detailed next.
Therefore, the only parameters that are learned during the mapping process are those of the hash grid $\Theta_i$.
The neural implicit map predicts a color $c$ and a signed distance function (SDF) value $s$ for any world coordinate $x$. 
To improve smoothness, the coordinate is first passed through a one-blob encoding $\gamma(\cdot)$ \cite{oneblob}. 
The geometry decoder $F_{\tau_i}$ then uses the encoded coordinate and features from the hash grid $V_{\Theta_i}(x)$ to output a feature vector $h$ and the SDF value: $ F_{\tau_i} \left( \gamma(x),  V_{\Theta_i}(x) \right) \to (h, s)$.
Finally, the color decoder $F_{\phi_i}$ uses this feature vector to output the RGB value $c$: $F_{\phi_i} \left( \gamma(x),  h) \right) \to c$.
To render RGB and depth images, UDON employs volumetric ray casting. 
For each pixel, we cast a ray and sample a sequence of points along it, query the color and SDF values from the neural implicit map, and integrate all the values along the ray to render the color and depth for the pixel.
As introduced in \eqref{eq:primal_update}, for every agent, we train a  neural implicit map $\Theta_i$ by minimizing a composite objective function, $L_i^{obj}$. For every agent $i$, this function combines several key loss terms: reconstruction losses for both color and depth against the ground truth images collected by the agent, a smoothness loss to encourage plausible geometry, and a free space loss that penalizes artifacts in areas far from object surfaces.

\subsection{Frequency-based Uncertainty} \label{method:uncertainty}
When merging a pair of neural implicit maps, $(\Theta_i, \Theta_j)$, having access to their respective uncertainties, $(u_i, u_j)$, is crucial. 
Different agents will have explored the same area to varying degrees, leading to local maps of differing quality and completeness.
The uncertainty allows the consensus optimization to fuse the two maps by selectively combining their most reliable portions. 
As a result, we can prevent fine details from being diluted or ``averaged out,'' leading to a more accurate reconstruction.
Motivated by this, we adopt the frequency-based uncertainty model from RAMEN \cite{ramen}. 
This model's key intuition is that gradient updates to vertices in the hash grid $V_{\Theta_i}$ are spatially localized, occurring only when an agent observes the corresponding area. 
This allows the number of gradient updates a vertex has received to serve as a direct proxy for its uncertainty, where a higher update count signifies lower uncertainty.
We compute the uncertainty of agent $i$'s map at iteration $t$, denoted as $u_i^t$:
\begin{equation}
    u_i^t = \sum_{k=0}^t \text{sgn} \left( \text{abs} \left( \frac{\partial  L^{obj}_i(\Theta_i^k, \mathcal{R}_i^k) }{\partial \Theta_i^k}  \right) \right).
    \label{eq:uncertainty}
\end{equation}
In \eqref{eq:uncertainty}, each element in the uncertainty vector $u_i^t$ simply tracks the number of times its corresponding hash grid parameter has received non-zero gradient updates.
The parts of the map that are not sampled during the training will correspond to the elements of $\Theta_i^k$ that have zero gradients.

To enable uncertainty-aware consensus between a pair of communicating agents $(i,j) \in \mathcal{E}$, we compute a pair of diagonal weight matrices, $W_{ij}^{t}$ and $W_{ji}^{t}$. 
These matrices are derived by normalizing the agents' respective uncertainty vectors, $u_i^{t}$ and $u_j^{t}$, into a user-specified range $[\beta_l, \beta_u]$ while preserving their relative proportions:
\begin{subequations}
    \label{eq:weight}
    \begin{gather}
        u_{sum} = u_i^{t} + u_j^{t}, \\
        \epsilon = \frac{\beta_{u} - \beta_{l}}{\text{max}(u_{sum}) - \text{min}(u_{sum}) } \label{eq:episilon}, \\
        \zeta = \beta_{l} - \epsilon \cdot \text{min}(u_{sum}) \label{eq:zeta}, \\
        W_{ij}^{t} = \text{diag}(\epsilon \cdot u_i^{t} + \zeta), \quad W_{ji}^{t} = \text{diag}(\epsilon \cdot u_j^{t} + \zeta).
    \end{gather}
\end{subequations}
This normalization relies on common scaling ($\epsilon$) and shifting ($\zeta$) factors computed from the combined uncertainty $u_{sum}$. 
This ensures that information from more certain maps (i.e., those with lower uncertainty values) is trusted more during the consensus process.
The resulting weight matrices are then incorporated into the uncertainty-weighted optimization, detailed in the next section.

\subsection{Uncertainty-weighted Distributed Optimization}
Our key innovation addresses an instability issue in the standard C-ADMM algorithm, introduced in Section \ref{Preliminaries}, by modifying the dual variables.
Prior C-ADMM based methods \cite{DiNNO,Di-NeRF,ramen} employ a \textbf{single, aggregate dual variable}, $p_i$, for each agent, as detailed in \eqref{eq:dual_update}. 
This variable \emph{accumulates consensus violations} from \emph{all} past neighbors. Consequently, agents that are no longer in communication ($j \notin \mathcal{N}_i^t$) continue to influence the optimization process. Under sparse or unreliable communication, this approach is problematic: an agent's map can be unduly penalized by outdated information from long-disconnected neighbors. This can cause the dual variable $p_i$ to grow without bound, leading to training instability or even divergence (which we will demonstrate in Section \ref{Experimental Results}).

In this work, we assign a unique dual variable $p_{(i,j)}$ to each communication link $(i,j) \in \mathcal{E}$. 
At each iteration $t$, this variable is updated only if the agents are actively communicating. 
The uncertainty-weighted dual update for communicating neighbors ($j \in \mathcal{N}_i^t$) is given by:
\begin{equation}
    p_{(i,j)}^{t+1} = p_{(i,j)}^t + 2 \rho W_{ij}^t (W_{ij}^t+W_{ji}^t)^{-1}(W_{ji}^t\Theta_i^{t+1} - W_{ji}^t\Theta_j^{t+1}),
    \label{eq:weight_dual}
\end{equation}
where the weight matrices $W_{ij}^t, W_{ji}^t$ are computed via \eqref{eq:weight}. 
For non-communicating neighbors ($j \notin \mathcal{N}_i^t$), the dual variable remains unchanged, $p_{(i,j)}^{t+1} = p_{(i,j)}^{t}$.
We will use the dual variable $p_{(i,j)}^{t+1}$ to penalize consensus violations in the following primal update.
For each agent, the primal update solves for new map parameters $\Theta_i^{t+1}$ by minimizing an objective that combines the local reconstruction loss with consensus terms from currently-active neighbors:
\begin{align}
    &\Theta_{i}^{t+1} = \argmin_{\Theta_i} \; L^{obj}_i(\Theta_i, \mathcal{R}_i^t) + \sum_{j \in \mathcal{N}_i^t}  \langle \Theta_i, p_{(i,j)}^{t+1} \rangle \nonumber \\
    & + \rho\sum_{j \in \mathcal{N}_i^t}  \big\Vert \Theta_i - (W_{ij}^t+W_{ji}^t)^{-1}(W_{ij}^t\Theta_i^t + W_{ji}^t\Theta_j^t) \big\Vert_{W_{ij}^t}^2,
    \label{eq:weight_primal_rewrite}
\end{align}
where $\Vert x \Vert^2_{W} = x^\top W x$ is a weighted norm.
The summations in \eqref{eq:weight_primal_rewrite} exclusively use information from currently-connected neighbors, which prevents the optimization from being corrupted by outdated data from long-disconnected agents. 
This approach, combined with uncertainty weighting, allows the consensus process to focus on reliable information from active peers, ensuring robust performance even under challenging communication conditions.
We designed Algorithm \ref{procedure_UDON} to be fully decentralized, where each agent is responsible for updating only its own dual variables and neural implicit map.

\definecolor{brightpurple}{HTML}{9900CC}
\definecolor{forestgreen}{HTML}{008631}

\begin{algorithm}
    \caption{UDON}
    \label{procedure_UDON}
    \begin{algorithmic}[1]

        \Statex \textbf{Initialization:}
        \For{$i \in \mathcal{V}$}
            \State $\Theta_i^0 = \Theta_{\text{initial}}$
            \For{$j \in \mathcal{V}, j \neq i$}
                \State $p_{(i,j)}^{0} = 0$ \Comment{Initialize all dual variables}
            \EndFor
        \EndFor

        \Statex
        \While {true} \Comment{Main Mapping Loop for Agent $i$}
            \State Send map $\Theta_i^t$ \& uncertainty $u_i^t$ to neighbors $\mathcal{N}_i^t$
            \State Receive $\Theta_j^t, u_j^t$ from neighbors $j \in \mathcal{N}_i^t$
            \State Compute weights $W_{ij}^{t}, W_{ji}^{t}$ for all $j \in \mathcal{N}_i^t$ using \eqref{eq:weight}

            \Statex
            
            \For{$j \in \mathcal{N}_i^t$} \Comment{\textcolor{brightpurple}{Dual Update}}
                 \State Update $p_{(i,j)}^{t+1}$ using dual update rule \eqref{eq:weight_dual}
            \EndFor

            \Statex
            
            \For{$b \leftarrow 1 \; \text{to} \; B$} \Comment{\textcolor{forestgreen}{Primal Update}}
                \State Sample a mini-batch from $\mathcal{R}_i^t$
                \State Update $\Theta_i$ via \eqref{eq:weight_primal_rewrite} to approximate $\Theta_{i}^{t+1}$
            \EndFor

            \Statex
            
            \State Compute uncertainty $u_i^{t+1}$ using \eqref{eq:uncertainty}

        \EndWhile
    \end{algorithmic}
\end{algorithm}

\section{EXPERIMENTAL RESULTS} \label{Experimental Results}

\begin{figure*}[hbt!]
	\centerline{\includegraphics[trim={0cm 0cm 0cm 0cm},clip,width=1.6\columnwidth]{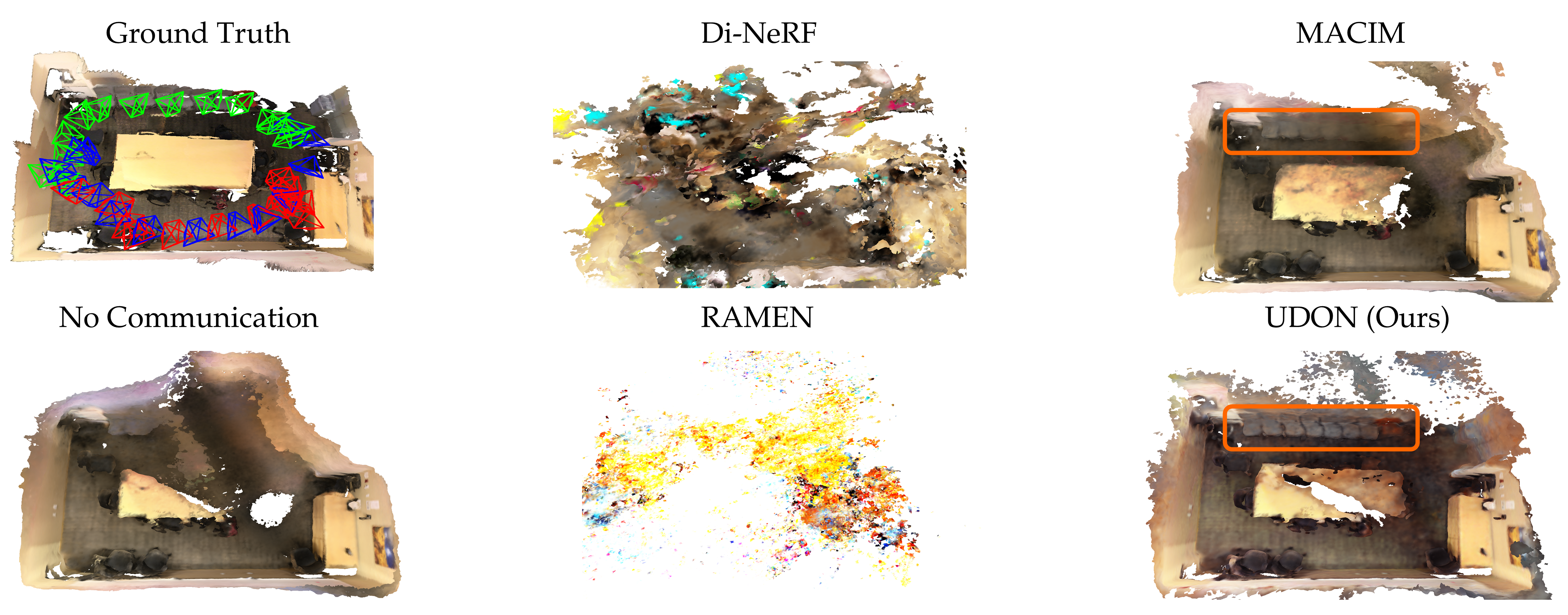}} 
	\caption{  \textbf{Qualitative comparison on ScanNet scene0169 with a challenging 1\% communication rate.} 
                The agent trajectories are color-coded on the ground truth mesh for reference. 
                Both Di-NeRF and RAMEN completely fail to reconstruct the scene. 
                In contrast, MACIM and UDON both produce coherent maps, demonstrating the significant benefit of even minimal communication over a no-communication baseline (map learned from only blue agent's observations). 
                However, a closer inspection (highlighted in orange) reveals UDON's superior ability to capture fine details, accurately reconstructing the chairs against the wall which MACIM misses. }
	\label{fig:exp1_vis}
\end{figure*}

\begin{table*}[!hbt]
  \caption{UDON outperforms the state-of-the-art methods at only 1\% communication success rate with three agents.}
  \centering
  \resizebox{1.7\columnwidth}{!}{%
    \begin{tabular}{cccccc}
      \toprule
      Method & Metric & office1 & room1 & scene0000 & scene0169 \\
      \midrule
      \multirow{3}{*}{No Communication}
        & \emph{Artifacts} $\downarrow$         & 13.76 $\pm$ 1.64 & 4.69 $\pm$ 2.36  & 17.25 $\pm$ 11.42  & 24.44 $\pm$ 6.23 \\ 
        & \emph{Holes} $\downarrow$       & 20.11 $\pm$ 4.94 & 6.90 $\pm$ 2.43 & 6.42 $\pm$ 2.67 & 6.30 $\pm$ 3.07 \\ 
        & \emph{Completion Ratio} (\%) $\uparrow$ & 51.64 $\pm$ 3.82 & 75.59 $\pm$ 8.73   & 71.80 $\pm$ 7.31   & 76.39 $\pm$ 7.17 \\ 
      \midrule
      \multirow{3}{*}{Di-NeRF~\cite{Di-NeRF}}
        & \emph{Artifacts} $\downarrow$         & 28.85$\pm$11.57 & 63.29$\pm$5.75  & 52.27$\pm$2.65  & 44.08 $\pm$ 9.72 \\ 
        & \emph{Holes} $\downarrow$       & 16.96$\pm$12.19 & 67.52$\pm$25.92 & 33.53$\pm$12.69 & 10.35 $\pm$ 1.69 \\ 
        & \emph{Completion Ratio} (\%) $\uparrow$ & 48.87$\pm$19.58 & 4.49$\pm$3.24   & 9.41$\pm$6.06   & 49.42 $\pm$ 5.26 \\ 
      \midrule
      \multirow{3}{*}{MACIM~\cite{MACIM}}
        & \emph{Artifacts} $\downarrow$         & 7.15$\pm$6.27   & 2.19$\pm$2.23   & 5.53$\pm$4.27   & 21.74 $\pm$ 6.99 \\ 
        & \emph{Holes} $\downarrow$       & 5.84$\pm$6.48   & 3.37$\pm$0.55   & 3.81$\pm$0.79   & 3.77 $\pm$ 1.13 \\
        & \emph{Completion Ratio} (\%) $\uparrow$ & 80.48$\pm$11.09 & 87.49$\pm$3.66  & 83.89$\pm$6.42  & 83.12 $\pm$ 4.78 \\
      \midrule
      \multirow{3}{*}{RAMEN~\cite{ramen}}
        & \emph{Artifacts} $\downarrow$         & 44.37$\pm$11.68 & 48.12$\pm$10.80 & 56.54$\pm$5.38  & 35.89 $\pm$ 3.37 \\
        & \emph{Holes} $\downarrow$       & 33.55$\pm$16.16 & 51.46$\pm$36.18 & 39.29$\pm$9.50  & 27.59 $\pm$ 7.13 \\
        & \emph{Completion Ratio} (\%) $\uparrow$ & 24.93$\pm$16.10 & 15.52$\pm$13.08 & 6.46$\pm$2.93   & 14.46 $\pm$ 6.62 \\
      \midrule
      \multirow{3}{*}{UDON (ours)}
        & \emph{Artifacts} $\downarrow$         & \color{DGreen}\textbf{4.40$\pm$3.27} & \color{DGreen}\textbf{1.37$\pm$0.05} & \color{DGreen}\textbf{2.83$\pm$0.20} & \color{DGreen}\textbf{20.83 $\pm$ 5.28} \\
        & \emph{Holes} $\downarrow$       & \color{DGreen}\textbf{3.44$\pm$0.90} & \color{DGreen}\textbf{3.11$\pm$0.08} & \color{DGreen}\textbf{3.14$\pm$0.26} & \color{DGreen}\textbf{3.30 $\pm$ 1.28} \\
        & \emph{Completion Ratio} (\%) $\uparrow$ & \color{DGreen}\textbf{84.35$\pm$6.76} & \color{DGreen}\textbf{89.46$\pm$0.24} & \color{DGreen}\textbf{90.67$\pm$2.89} & \color{DGreen}\textbf{85.38 $\pm$ 6.13} \\
      \bottomrule
    \end{tabular}%
  }
  \label{tab:simtest_1}
\end{table*}

In this section, we present a comprehensive evaluation of our method with experiments designed to answer the following key questions:
\begin{enumerate}
    \item How does UDON perform across a range of communication success rates, from moderate (20\%) down to extremely poor (1\%)?
    \item How does UDON's performance scale as the number of agents increases from 4 to 8?
    \item How effectively does UDON handle real-world data collected by robotic platforms?
    \item What enables UDON's high reconstruction performance when other C-ADMM-based methods fail under severe communication disruptions?
\end{enumerate}
To answer these questions, we evaluate UDON's performance on standard synthetic and real-world benchmarks.
Specifically, we used the Replica dataset \cite{replica19arxiv} for its synthetic indoor scenes with posed RGB-D images and the ScanNet dataset \cite{dai2017scannet} for its challenging real-world RGB-D scans of large indoor environments. 
We also collected a custom hardware dataset using TurtleBot robots.
We compared UDON against state-of-the-art multi-agent neural implicit mapping methods, including Di-NeRF \cite{Di-NeRF}, MACIM \cite{MACIM}, and RAMEN \cite{ramen}. 
For fair comparison, all methods adopt the same neural implicit representation, detailed in section \ref{method:mapping}.
We conducted our experiments on an RTX 5090 GPU. 
For a single agent, UDON runs at 5.3 FPS and occupies approximately 3.8 GB of GPU memory. 

In our setup, each agent attempts to communicate with other agents at every mapping iteration. 
However, we simulate challenging conditions by setting a low \emph{communication success rate (\%)}. 
For example, a 1\% rate means that, on average, an agent receives a successful update from any given neighbor only once every 100 iterations, forcing it to rely heavily on outdated information.
To account for the stochastic nature of communication dropouts, we conduct each experiment across three independent trials and report the mean and variance of the performance metrics.

\begin{figure*}[h!]
	\centerline{\includegraphics[trim={0cm 0cm 0cm 0cm},clip,width=1.5\columnwidth]{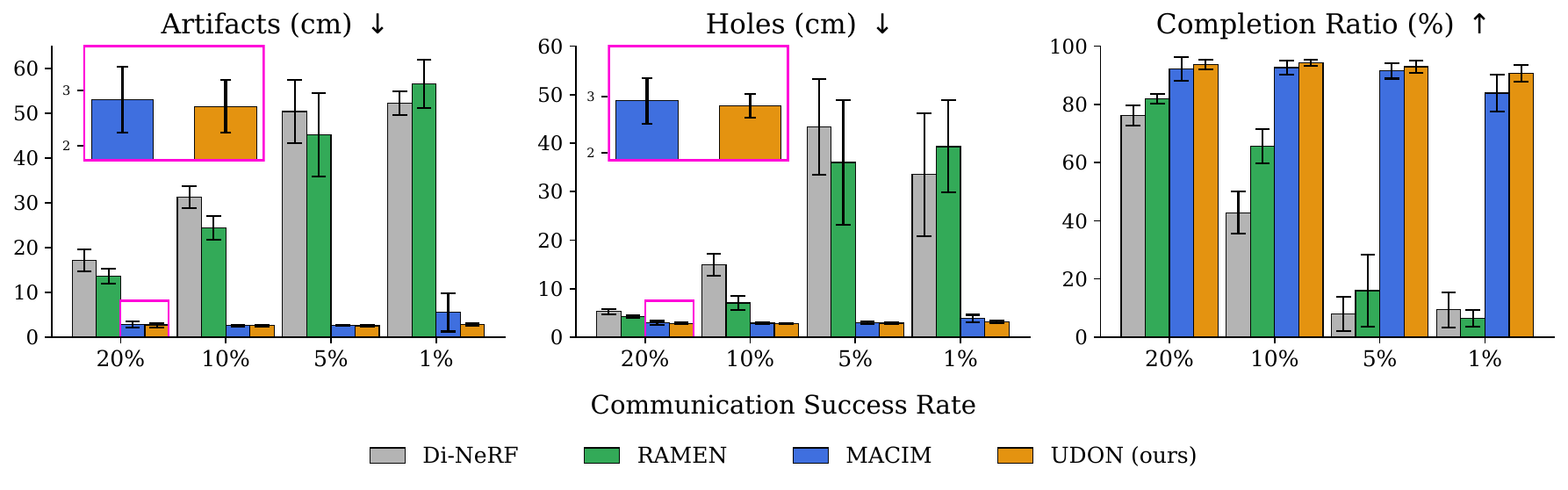}} 
	\caption{
    \textbf{Comparison of reconstruction quality on scene0000 across four communication success rates (1\%, 5\%, 10\%, 20\%)}.
            UDON (ours) achieves the lowest artifacts and holes and the highest completion ratio, with small variance. 
            MACIM performs reasonably but falls behind UDON as communication degrades, while Di-NeRF and RAMEN collapse. 
            These results highlight UDON’s robustness under constrained communication.}
	\label{fig:bar}
\end{figure*}

\begin{figure*}
	\centerline{\includegraphics[trim={0cm 0cm 0cm 0cm},clip,width=1.5\columnwidth]{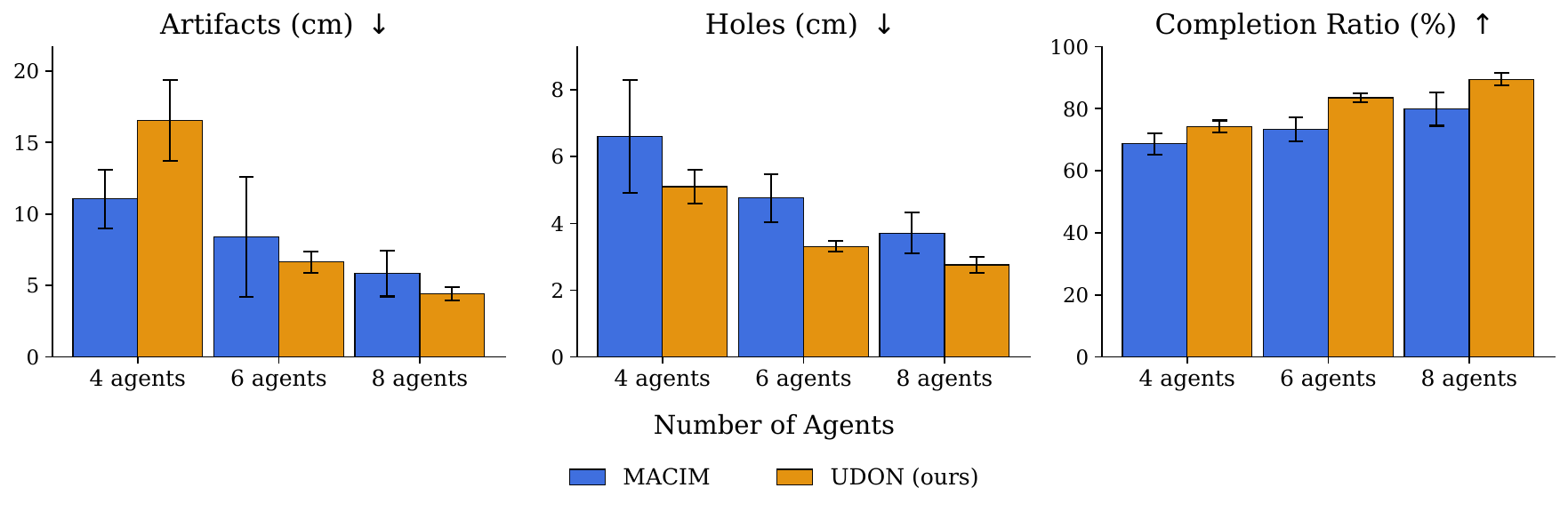}} 
	\caption{
    \textbf{Comparison of reconstruction quality on scene0000 with varying numbers of communicating agents (4, 6, 8) at 5\% communication success rate.} 
    UDON (ours) consistently outperforms MACIM. 
    Furthermore, reconstruction quality improves as the number of participating agents increases, 
    indicating that even under low communication rates, leveraging more agents 
    provides clear benefits.
}
	\label{fig:bar2} 
\end{figure*}

To evaluate geometric accuracy, we first generated meshes using the marching cubes algorithm \cite{marching_cube}. 
We adopted three metrics to assess reconstruction quality against the ground truth:
\begin{enumerate}
\item \emph{Artifacts} (cm): The average distance from the reconstructed mesh to the nearest ground truth, quantifying the amount of reconstructed artifacts. Lower is better. 
\item \emph{Holes} (cm) \footnote{The \emph{Holes} metric differs from \emph{Artifacts} because it does not penalize disconnected artifacts (i.e., ``floaters'') that are far from the actual surface.}
: The average distance from the ground truth to the nearest reconstructed mesh, measuring completeness. Lower is better.

\item \emph{Completion Ratio} (\%): The percentage of the ground truth surface reconstructed within a 5 cm tolerance. Higher is better.
\end{enumerate}

\subsection{Replica and ScanNet Datasets}
\textbf{How does UDON perform under extremely low communication success rates (1\%)}?
We evaluated  UDON against Di-NeRF, MACIM, and RAMEN on two Replica scenes (office1, room1) and two ScanNet scenes (scene0000, scene0169), under a  setting with three agents and a communication success rate of 1\%.
The trajectories of the three agents are color-coded in Fig. \ref{fig:exp1_vis}.
This scenario simulates the severe communication constraints faced by resource-constrained robotic teams in isolated environments, such as during subterranean exploration. 
The results, presented in Table \ref{tab:simtest_1} and Fig. \ref{fig:exp1_vis}, demonstrate UDON's superior performance across all metrics. While MACIM is competitive on simpler scenes, its quality degrades on the more complex ScanNet dataset, where UDON remains consistently strong and better captures fine details (highlighted in orange in Fig. \ref{fig:exp1_vis}). 
In contrast, both RAMEN and Di-NeRF completely fail to reconstruct coherent scenes under these conditions. 
Furthermore, UDON exhibits consistently smaller standard deviations, evidencing a more stable and robust pipeline. 
These findings also highlight a key point: \emph{even a minimal 1\% communication rate allows for the exchange of critical information, leading to significantly more complete reconstructions compared to a no-communication baseline (map learned from one agent's observations)}.

\textbf{How does UDON perform across a range of different communication rates (20\%, 10\%, 5\%, 1\%)?} 
We show the various communication rate results in Fig. \ref{fig:bar}.
UDON demonstrates strong resilience across a wide range of communication rates. 
At moderate rates (10-20\%), it achieves near-optimal reconstruction quality, with minimal artifacts and completion ratios above 90\%. 
Crucially, as the rate drops to 5\% and even 1\%, UDON's performance degrades gracefully, maintaining stable variance while outperforming all baselines. 
In contrast, while Di-NeRF and RAMEN perform reasonably at higher rates, they fail completely under the more extreme conditions (1-5\%). 
MACIM remains competitive but suffers a sharp performance decline when the communication rate drops from 5\% to 1\%. 
These results highlight the effectiveness of UDON in sustaining accurate and consistent mapping, allowing it to excel even under the most severe communication constraints.
\begin{figure*}[hbt!]
	\centerline{\includegraphics[trim={3cm 2cm 3cm 1.5cm},clip,width=1.7\columnwidth]{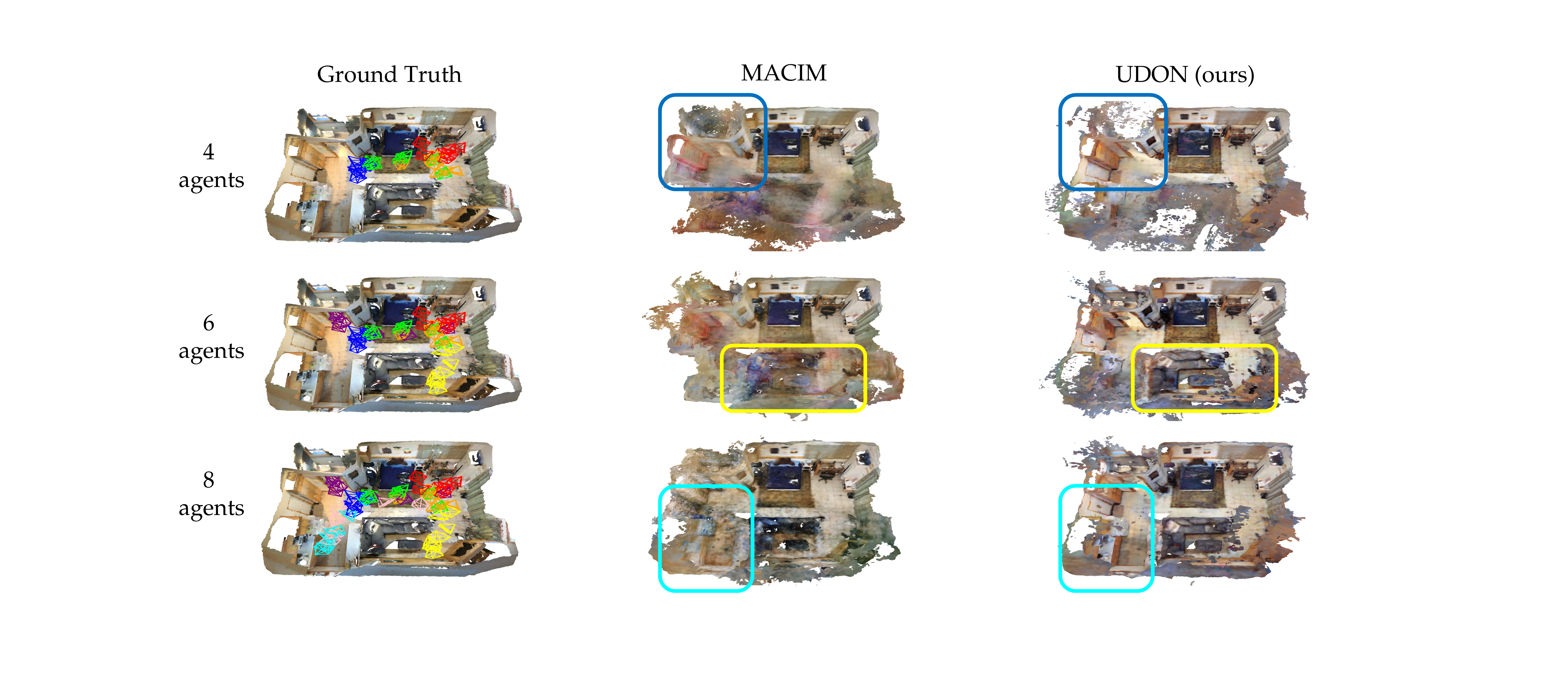}} 
	\caption{  \textbf{Qualitative comparison on ScanNet scene0000 with an increasing number of agents (4, 6, 8 agents) at 5\% communication success rate.}
    As more agents are added, UDON effectively leverages the additional data to produce increasingly complete scene reconstructions. 
    Compared to MACIM, UDON demonstrates a significant advantage in areas observed by only a few agents (highlighted regions, color-coded by the corresponding agent's trajectory). 
    Furthermore, UDON captures color and texture with higher fidelity as evidenced by the accurately-reconstructed floor tiles.}
	\label{fig:exp3_vis}
\end{figure*}

\begin{figure*}[hbt!]
	\centerline{\includegraphics[trim={0cm 0cm 0cm 0cm},clip,width=1.5\columnwidth]{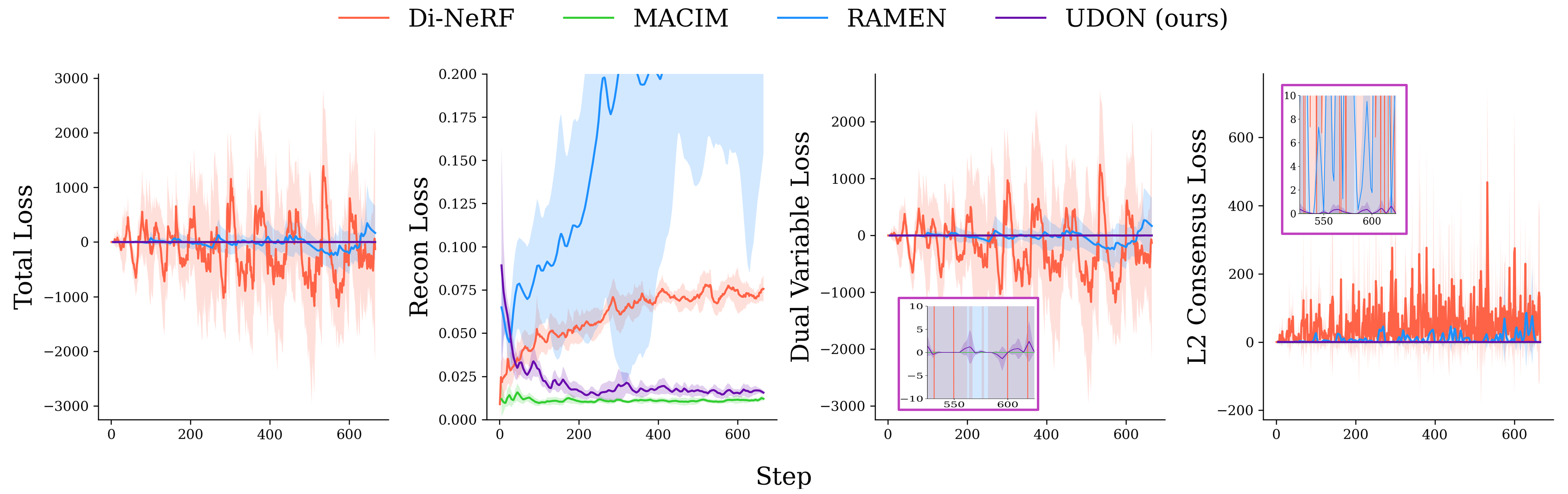}} 
	\caption{
    \textbf{Analysis of the primal update loss curves for different methods.} The total loss, defined in \eqref{eq:weight_primal_rewrite}, is composed of the \emph{Reconstruction Loss}, \emph{Dual Variable Loss} (consensus enforcement), and \emph{L2 Consensus Loss} (quadratic penalty). The dual variable loss for Di-NeRF and the RAMEN diverges, causing training instability and degrading reconstruction quality. In contrast, UDON stabilizes this loss term. This stability allows UDON to achieve a consistently low reconstruction loss.}
	\label{fig:loss}
\end{figure*}

\textbf{How does UDON perform with an increasing number of agents (4, 6, 8 agents)?} 
We further analyze how performance scales with the number of agents on scene0000 at 5\% communication success rate. 
The quantitative results in Fig. \ref{fig:bar2} show that UDON consistently outperforms MACIM.
UDON exhibits a higher artifact score only in the 4-agent setting. 
As shown in Fig. \ref{fig:exp3_vis}, this is primarily caused by ``floaters'' appearing in completely unobserved regions of the scene.
Notably, UDON's reconstruction quality improves markedly as the agent count increases from 4 to 8, demonstrating its ability to effectively integrate complementary information from additional teammates even with sparse communication. 
The visual results in Fig. \ref{fig:exp3_vis} support these findings. 
Compared to MACIM, UDON maintains high fidelity in regions with sparse coverage from only a few agents (highlighted areas in dark blue, yellow, and cyan for the corresponding color-coded agents).
This advantage is enabled by UDON's uncertainty-aware optimization. 
This mechanism prevents high-quality observations from being ``averaged out'' by less-certain information. 
UDON also produces more photorealistic textures, as evidenced by the detailed rendering of the floor tiles.

\textbf{How does UDON avoid mapping failures under extreme communications?}
The effectiveness of UDON stems from assigning an unique dual variable for each communication pair, which stabilizes the training process, as we will illustrate by analyzing the loss curves in Fig. \ref{fig:loss}.
To interpret the figure, we first define the components of the \emph{Total Loss} in \eqref{eq:weight_primal_rewrite}. 
The \emph{Reconstruction Loss} refers to $L^{obj}_i(\Theta_i, \mathcal{R}_i^t)$. 
The consensus objective is composed of two parts: the \emph{Dual Variable Loss}, which for UDON is $\sum_{j \in \mathcal{N}_i} \langle \Theta_i, p_{(i,j)}^{t+1} \rangle$, and the \emph{L2 Consensus Loss}, which is the final quadratic penalty term in the equation.
Fig. \ref{fig:loss} reveals that both Di-NeRF and the RAMEN suffer from training instability. 
Their \emph{Dual Variable Loss} oscillates with growing magnitude, causing the total loss to diverge and the reconstruction quality to degrade. 
While MACIM avoids this issue by removing dual variables from its optimization, this can limit its ability to capture fine scene details, a limitation observed in our previous experiments.
In contrast, UDON provides a robust alternative. 
By ensuring that only active communication links contribute to the \emph{Dual Variable Loss}, it prevents its cost term from growing excessively.

\subsection{Turtlebot Hardware Dataset}
We further evaluate UDON's performance on a dataset collected using TurtleBot robots. 
This dataset presents a more realistic challenge than standard benchmarks like Replica and ScanNet, which are captured with hand-held cameras.
Due to the TurtleBots' low platform height and constrained mobility, imaging is frequently obstructed by obstacles, resulting in highly-incomplete views for any single agent. 
This scenario rigorously tests the multi-robot team's ability to effectively fuse shared information to complete the scene from partially-obstructed viewpoints.
To form our experiment, we collected three scenes and segmented the data to simulate a three-agent scenario with a challenging 5\% communication success rate.
We recorded ground truth camera poses using a VICON motion capture system, and we enhanced depth images via the model in \cite{liu2025manipulation}.
As shown in Table \ref{tab:hardware} and Figure \ref{fig:teaser}, UDON significantly outperforms the baseline methods in this realistic setting. 
Notably, the performance gap between UDON and MACIM is even larger here than in the previous experiments (Table \ref{tab:simtest_1}). 
This result further demonstrates UDON's robust capability to handle the complexities of real-world multi-robot mapping under severe communication constraints.

\vspace{2mm}
\begin{table}[!hbt]
  \caption{Performance Comparison on the TurtleBot Dataset.}
  \label{tab:hardware}
  \centering
  \resizebox{\columnwidth}{!}{%
  \begin{tabular}{ccccc}
    \toprule
    Method & Metric & exp1 & exp2 & exp3 \\
    \midrule
    \multirow{3}{*}{MACIM~\cite{MACIM}}
      & \emph{Artifacts} $\downarrow$                                   & 6.91 $\pm$ 0.80 & 5.92 $\pm$ 0.70 & 5.64 $\pm$ 1.25 \\
      & \emph{Holes} $\downarrow$                                       & 5.64 $\pm$ 0.80 & 4.36 $\pm$ 0.16 & 5.61 $\pm$ 0.38 \\
      & \makecell[l]{\emph{Completion}\\\emph{Ratio} (\%) $\uparrow$}   & 66.52 $\pm$ 3.54 & 75.35 $\pm$ 0.67 & 65.69 $\pm$ 3.16 \\
    \midrule
    \multirow{3}{*}{UDON (ours)}
      & \emph{Artifacts} $\downarrow$                                   & \color{DGreen}\textbf{5.24 $\pm$ 0.68}  & \color{DGreen}\textbf{4.86 $\pm$ 0.58}  & \color{DGreen}\textbf{4.58 $\pm$ 0.42} \\
      & \emph{Holes} $\downarrow$                                       & \color{DGreen}\textbf{3.92 $\pm$ 0.58}  & \color{DGreen}\textbf{3.29 $\pm$ 0.38}  & \color{DGreen}\textbf{3.60 $\pm$ 0.38} \\
      & \makecell[l]{\emph{Completion}\\\emph{Ratio} (\%) $\uparrow$}   & \color{DGreen}\textbf{78.75 $\pm$ 4.41} & \color{DGreen}\textbf{83.64 $\pm$ 3.15} & \color{DGreen}\textbf{80.21 $\pm$ 3.35} \\
    \bottomrule
  \end{tabular}%
  }
\end{table}

\section{CONCLUSIONS}\label{Conclusion}
In this paper, we introduced UDON, a novel uncertainty-weighted distributed optimization method for multi-robot neural implicit mapping, specifically designed for teams operating under severe communication constraints. 
Our comprehensive experiments demonstrate that UDON successfully reconstructs complete scenes with fine details, even at extremely low communication rates. 
Furthermore, our results show that the framework scales effectively, improving reconstruction quality as more agents are added to the team.
For future work, extending UDON to the mapping of large-scale environments by incorporating gradually-added submaps is a promising direction.

\section*{Acknowledgments}
This work was supported in part by NSF grants ECCS-2438314 (CAREER Award) and CNS-2529645.
Additionally, this research was made possible by GPU resources provided via the NVIDIA Academic Grant.








\bibliographystyle{IEEEtran} 
\bibliography{IEEEabrv, root} 


\end{document}